\documentclass[10pt,twocolumn,letterpaper]{article}
\usepackage{cvpr}
\usepackage[]{authblk}
\usepackage{times}
\usepackage{epsfig}
\usepackage{graphicx}
\usepackage{amsmath}
\usepackage{amssymb}
\usepackage{bm}
\usepackage{multirow}
\usepackage{algorithm}
\usepackage{algorithmic}
\usepackage[tight,footnotesize]{subfigure}
\renewcommand{\algorithmicrequire}{\textbf{Input:}} 
\renewcommand{\algorithmicensure}{\textbf{Iteration:}} 
\renewcommand{\algorithmiclastcon}{\textbf{Output:}}

\usepackage[breaklinks=true,bookmarks=false]{hyperref}

\cvprfinalcopy 

\def\cvprPaperID{****} 

\begin{document}

\title{Bayes Merging of Multiple Vocabularies for Scalable Image Retrieval}

\author[1]{Liang Zheng}
\author[1]{Shengjin Wang}
\author[2]{Wengang Zhou}
\author[3]{Qi Tian}
\setlength{\affilsep}{0em}\affil[1]{State Key Laboratory of Intelligent Technology and Systems;} \affil[1]{Tsinghua National Laboratory for Information Science and Technology;}\affil[1]{Department of Electronic Engineering, Tsinghua University, Beijing 100084, China}
\affil[2]{EEIS Department, University of Science and Technology of China}
\affil[3]{University of Texas at San Antonio, TX, 78249, USA
\authorcr{\tt\small zheng-l06@mails.tsinghua.edu.cn \tt\small wgsgj@tsinghua.edu.cn \tt\small zhwg@ustc.edu.cn \tt\small qitian@cs.utsa.edu}
}
\maketitle

\begin{abstract}
In the Bag-of-Words (BoW) model,
 the vocabulary is of key importance. Typically, multiple vocabularies are generated to correct quantization artifacts and improve recall. However, this routine is corrupted by vocabulary correlation, i.e., overlapping among different vocabularies. Vocabulary correlation leads to an over-counting of the indexed features in the overlapped area, or the intersection set, thus compromising the retrieval accuracy. In order to address the correlation problem while preserve the benefit of high recall, this paper proposes a Bayes merging approach to down-weight the indexed features in the intersection set. Through explicitly modeling the correlation problem in a probabilistic view, a joint similarity on both image- and feature-level is estimated for the indexed features in the intersection set.

We evaluate our method on three benchmark datasets. Albeit simple, Bayes merging can be well applied in various merging tasks, and consistently improves the baselines on multi-vocabulary merging. Moreover, Bayes merging is efficient in terms of both time and memory cost, and yields competitive performance with the state-of-the-art methods.
\end{abstract}

\section{Introduction}
This paper considers the task of Bag-of-Words (BoW) based image retrieval, especially on multi-vocabulary merging. We aim at improving the retrieval accuracy while maintaining affordable memory and time cost.


\begin{figure}
  \centering
  \includegraphics[width=2.9in]{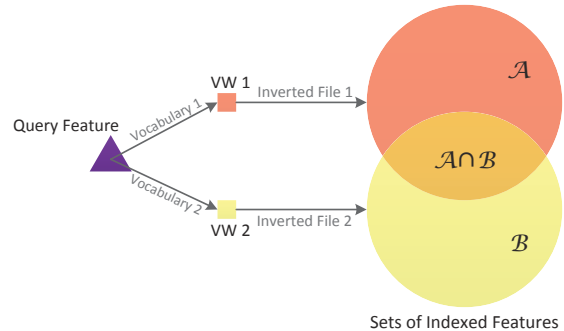}\\
  \caption{An Illustration of vocabulary correlation. Given a query feature, it is quantized to two visual words in two vocabularies. Then, two sets of indexed features, $\mathcal{A}$ and $\mathcal{B}$, are identified from the two inverted files, respectively. The area of the intersection set $\mathcal{A}\cap \mathcal{B}$ (denoted as $Card(\mathcal{A}\cap \mathcal{B})$) encodes the extent of correlation between the two sets. In this paper, we focus on the indexed features in $\mathcal{A}\cap \mathcal{B}$. }\label{fig:intersection}
\end{figure}

The \emph{vocabulary} (also called the codebook or quantizer) lies at the core of the BoW based image retrieval system. It functions by quantizing SIFT descriptors \cite{SIFT2} to discrete \emph{visual words}. The quantized visual words are the nearest centers to the feature vectors in the feature space. In order to reduce quantization error and improve recall, multiple vocabularies are often generated, and each feature is quantized to different visual words from multiple vocabularies. The primary benefit of using multiple vocabularies is that more candidate features are recalled, which corrects quantization artifacts to some extent.

However, the routine of multi-vocabulary merging is affected by a crucial problem, \ie, vocabulary correlation \cite{jegou2012negative} (see Fig. \ref{fig:intersection}). Given a query feature, based on the inverted files with two individual vocabularies, two sets of indexed features $\mathcal{A}$ and $\mathcal{B}$ are identified, sharing an intersection set $\mathcal{A}\cap \mathcal{B}$. In this paper, the area of $\mathcal{A}\cap \mathcal{B}$ is approximated by $Card (\mathcal{A}\cap \mathcal{B})$. The larger $Card(\mathcal{A}\cap \mathcal{B})$ is, the larger the correlation will be. In an extreme case, total correlation occurs if $Card(\mathcal{A}\cap \mathcal{B}) = Card(\mathcal{A}\cup \mathcal{B})$, and merging $\mathcal{A}$ and $\mathcal{B}$ brings no benefit.

A straightforward method for multi-vocabulary merging consists in concatenating the BoW histograms of different vocabularies \cite{HKM}. In a microscopic view of this method, the indexed features in $\mathcal{A}\cap \mathcal{B}$ are counted twice in Fig. \ref{fig:intersection}. Nevertheless, since images in this area are mostly irrelevant ones (the number of relevant images is always very small), the over-counting may actually compromise the retrieval accuracy \cite{AKM}.

In this paper, we consider the situation in which the given vocabularies are correlated, and we aim to reduce the impact of correlation. To address this problem, this paper proposes to model the vocabulary correlation problem from a probabilistic view. In a nutshell, we jointly estimate an image- and feature-level similarity for the indexed features in the intersection set (or overlapping area). Given a query feature, lists of indexed features are extracted from multiple inverted files. Then, we identify the intersection and union sets of the lists, from which the cardinality ratio is calculated. This ratio thus encodes the extent of correlation (see Fig. \ref{fig:intersection}). For the indexed images in the intersection set, its similarity with the query is estimated as a function of the cardinality ratio, and subsequently added to the matching score. Experiments on several benchmark datasets demonstrate that Bayes merging is effective, and yields competitive results with the state-of-the-art methods.



\section{Related Work}
\label{section: related_work}
\textbf{Vocabulary Generation} The vocabulary provides a discrete partitioning of the feature space by visual words. Typically, either flat kmeans \cite{AKM, hamming} or hierarchical kmeans \cite{HKM} is employed to train a vocabulary in an unsupervised manner.
Improved methods include incorporating contextual information into the vocabulary \cite{contextual_visual_vocabulary}, building super-sized vocabulary \cite{contextual_weighting, co_indexing, mikulik2010learning}, making use of the active points \cite{wang2012fast}, etc.

\textbf{Matching Refinement} Feature-to-feature matching is a key issue in the BoW model. The baseline approach employs a coarse word-to-word matching, resulting in undesirable low precision. To improve precision, some works analyze the spatial contexts \cite{contextual_weighting, zheng2013visual, zhou2012principal} of SIFT features, and use the spatial constraints as solution to refining matching. Another line of works extracts binary signatures from SIFT descriptors \cite{hamming} or its contexts \cite{cMI, liu2014contextual}. The feature matching is thus refined by a further check of the Hamming distance between binary signatures. In this paper, however, we argue that even if two features are adjacent in the feature space, the corresponding images are probably very different. Therefore, we are supposed to look one step further by estimating a joint similarity on both image- and feature-level from clues in multiple vocabularies.

\textbf{Multiple Vocabularies}
It is well known that multi-vocabulary merging is effective in improving recall \cite{jegou2012negative, joint_index}. Typically, multi-vocabulary merging can be performed either at score level, \eg, by concatenating the BoW histograms \cite{HKM}, or at rank level, \eg, by rank aggregation \cite{jegou2010accurate}. On the other hand, some works also provide clues that multiple vocabularies also improve precision \cite{babenko2012inverted, wu2009multi}.
To address the problem of vocabulary correlation, Xia \etal \cite{joint_index} propose to create the vocabularies jointly and reduce correlation from the view of vocabulary generation. A more relevant work includes \cite{jegou2012negative}, which uses PCA to implicitly remove correlation of given vocabularies, resulting in a low dimensional image representation. Our work departs from previous works in two aspects. First, we explicitly model the vocabulary correlation problem from a probabilistic view. Second, our work is proposed for the BoW based image retrieval task, which differs from NN search problems.

\section{Background}
\label{section: background_baseline}
\subsection{Notations}
Assume that the $K$ vocabularies are denoted as  $\mathcal{V}^{(k)} = \{v_1^{(k)}, v_2^{(k)}, ..., v_{s_k}^{(k)}\}, k = 1,...,K$, where $v_i^{(k)}$ represents a visual word and $s_k$ is the vocabulary size. Correspondingly, built on $\mathcal{V}^{(k)}$, $K$ inverted files are organized as $\mathcal{W}^{(k)} = \{W_1^{(k)}, W_2^{(k)}, ..., W_{s_k}^{(k)}\}, k = 1,...,K$, where each entry $W_i^{(k)}$ contains a list of indexed features.

Given a query SIFT feature $x$, it is quantized to a visual word tuple $\left(v^{(1)}, v^{(2)},...,v^{(K)}\right)$, where $v^{(k)}, k=1,...,K$ is the nearest centroid in $\mathcal{V}^{(k)}$ to $x$. With the $K$ visual words we can identify $K$ sets of indexed features in entries $\{W_{i_k}^{(k)}\}_{k=1}^K$.
From the $K$ sets, we can define three types of sets to be used in this paper.
\newtheorem{definition}{Definition}
\begin{definition}[$\bm k^{th}$-order intersection set]
The intersection set of $k$, and only $k$ sets, denoted as $\cap^{(k)}$, $k\geq2$.
\end{definition}
\begin{definition}[$\bm k^{th}$-order union set]
The union set of $k$, and only $k$ sets, denoted as $\cup^{(k)}$, $k\geq2$.
\end{definition}
\begin{definition}[difference set]
The set in which no overlapping exists, i.e., $\cup^{(K)} - \sum_{k = 2}^{K} \cap^{(k)}$, $K\geq2$.
\end{definition}

\subsection{Baselines}
\label{section: baseline}

\textbf{Single vocabulary baseline} (B$_0$)
For a single vocabulary, we adopt the baseline introduced in \cite{AKM, hamming}. Specifically, vocabularies are trained by AKM on the independent Flickr60K data \cite{hamming}, and average IDF \cite{zheng2013lp} weighting scheme is used. We replace the original SIFT descriptor with rootSIFT \cite{root_sift}. In this scenario, we denote the matching function between two features $x$ and $y$ as,
\begin{equation}\label{equation:matching_1vocab}
  f_0(x, y) = \delta_{v_x, v_y}
\end{equation}
where $v_x$ and $v_y$ are visual words of $x$ and $y$ in the vocabulary, respectively, and $\delta(\cdot)$ is the Kronecker delta response.

\textbf{Conventional vocabulary merging} (B$_1$) Given $K$ vocabularies, B$_1$ simply concatenates multiple BoW histograms \cite{HKM}. It is equivalent to a simple score-level addition of the outputs of multiple vocabularies. The matching function between features $x$ and $y$ can be defined as

\begin{equation}\label{equation:matching_2vocab}
  f_1(x, y) = \sum_{k = 1}^{K}\delta_{v_x^{(k)}, v_y^{(k)}}
\end{equation}
where $v_x^{(k)}$ and $v_y^{(k)}$ are visual words in vocabulary $\mathcal{V}^{(k)}$ for $x$ and $y$, respectively. Eq. \ref{equation:matching_2vocab} shows that in baseline B$_1$, an indexed feature is counted $k$ times if it is in the $k^{th}$-order intersection set $\cap^{(k)}$, and only once if in the difference set (since there is no overlapping).

\textbf{Multi-index based vocabulary merging} (B$_2$) In \cite{babenko2012inverted}, a multi-index is organized as a multi-dimensional structure. In its nature, given $K$ vocabularies, two features are considered as a match \emph{iff} they are in the $K^{th}$-order intersection set $\cap^{(K)}$ of the indexed feature lists. Therefore, in baseline B$_2$, the matching function is defined as
\begin{equation}\label{equation:matching_multi_index}
  f_2(x, y) = \prod_{k=1}^{K}\delta_{v_x^{(k)}, v_y^{(k)}}
\end{equation}
Eq. \ref{equation:matching_multi_index} only counts the indexed features in $\cap^{(K)}$, discarding the rest. Therefore, the recall is low for B$_2$.

\section{Proposed Method}
\label{section: proposed_method}
For multi-vocabulary merging, the major problem is the over-counting of the intersection sets $\cap^{(k)}, k = 1,...,K$. On the other hand, the major benefit is a high recall, which is encoded in the difference set. Taking both issues into consideration, we propose to exert a likelihood on the intersection sets and preserve the difference set (scored as B$_1$). Without loss of generality, we start from the case of two vocabularies and then generalize it to multiple vocabularies.

\begin{figure}
  \centering
  \includegraphics[width=3.25 in]{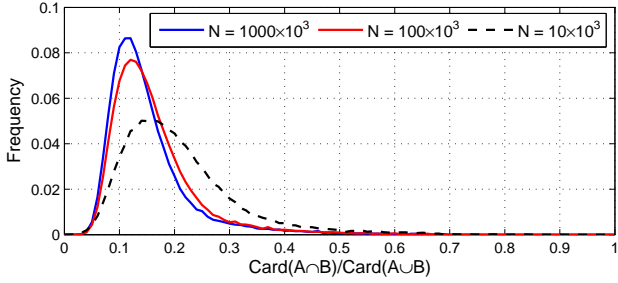}\\
  \caption{The distribution of the cardinality ratio on 10K, 100K, 1000K images, respectively. We use two vocabularies of size 20K.}\label{fig:inter_area_distribution}
\end{figure}

\subsection{Model Formulation}
\label{section: model}
Given a query feature $x$ in image $Q$, two sets of indexed features $\mathcal{A}$ and $\mathcal{B}$ are identified in two inverted files, respectively. Here, we want to evaluate the likelihood that a SIFT feature $y$ is a true neighbor of $x$ given that $y$ belongs to the intersection set of $\mathcal{A}$ and $\mathcal{B}$. This likelihood can be modeled as the following conditional probability,
\begin{equation}\label{equation: conditional_probability}
  w(x, y) = p(y \in T_x \left.\right| y \in  \mathcal{A} \cap \mathcal{B}).
\end{equation}
In Eq. \ref{equation: conditional_probability}, we define $T_x$ as the set of features which are visually similar to $x$ (locally) \emph{and} belong to the ground truth images of $Q$ (globally). On the other hand, $F_x$ is defined as the features which violate any of the two criteria. Therefore, $T_x$ and $F_x$ satisfy the follows
 \begin{equation}\label{equation: Tx_Fx}
   p(y\in T_x) + p(y \in F_x) = 1.
 \end{equation}
 For simplicity, we denote $y \in  \mathcal{A} \cap \mathcal{B}$ as $\mathcal{A} \cap \mathcal{B}$, $y \in T_x$ as $T_x$, and $y \in T_y$ as $T_y$, Then, using the formula of Bayes' theorem as well as Eq. \ref{equation: Tx_Fx}, we get
\begin{equation}\label{equation: Bayes}
\begin{aligned}
  p(T_x &\left.\right| \mathcal{A} \cap \mathcal{B}) = \frac{p( \mathcal{A} \cap \mathcal{B}\left.\right| T_x)\cdot p( T_x)}{p( \mathcal{A} \cap \mathcal{B})}\\
  &= \frac{p( \mathcal{A} \cap \mathcal{B}\left.\right| T_x)\cdot p(T_x)} {p( \mathcal{A} \cap \mathcal{B} \left.\right| T_x)\cdot p(T_x) + p(\mathcal{A} \cap \mathcal{B} \left.\right| F_x)\cdot p(F_x)}.
  \end{aligned}
\end{equation}
Then, re-formulating Eq. \ref{equation: Bayes}, we have
\begin{equation}\label{equation:Bayes_reform}
p(T_x \left.\right| \mathcal{A} \cap \mathcal{B}) = \left(1 + \frac{p( \mathcal{A} \cap \mathcal{B}\left.\right| F_x)}{p( \mathcal{A} \cap \mathcal{B}\left.\right| T_x)}\cdot\frac{p(F_x)}{p(T_x)}\right)^{-1}.
\end{equation}
In Eq. \ref{equation:Bayes_reform}, there are actually three random variables to estimate, \ie, $p( \mathcal{A} \cap \mathcal{B}\left.\right| F_x)$ (\textbf{term 1}), $p( \mathcal{A} \cap \mathcal{B}\left.\right| T_x)$ (\textbf{term 2}),  and $p(F_x) \slash p(T_x)$ (\textbf{term 3}). In Section \ref{section: prob_estimation}, we will exploit the estimation of these probabilities.

\subsection{Probability Estimation}
\label{section: prob_estimation}
\textbf{Estimation of term 1} In Eq. \ref{equation:Bayes_reform}, the term $p( \mathcal{A} \cap \mathcal{B}\left.\right| F_x)$ encodes the probability that feature $y$ lies in the set $\mathcal{A} \cap \mathcal{B}$ given that $y$ is a false match of query feature $x$. In this case, we should consider the distribution of the $x$' false matches in sets $\mathcal{A}$ and $\mathcal{B}$. In large databases, the number of true matches (both locally and globally) is limited. In other words, false matches dominate the space covered by $\mathcal{A}$ and $\mathcal{B}$. Therefore, we assume that false matches are uniformly distributed in $\mathcal{A}$ and $\mathcal{B}$, and term 1 can be estimated as
\begin{equation}\label{equation: term1}
  p( \mathcal{A} \cap \mathcal{B}\left.\right| F_x) = \frac{Card(\mathcal{A} \cap \mathcal{B})}{Card(\mathcal{A} \cup \mathcal{B})},
\end{equation}
where $Card(\cdot)$ represents the cardinality of a set. Eq. \ref{equation: term1} implies that, the probability that a false match falls into $\mathcal{A}\cap \mathcal{B}$ is proportional to the cardinality ratio $\frac{Card(\mathcal{A} \cap \mathcal{B})}{Card(\mathcal{A}\cup \mathcal{B})}$. Intuitively, the larger the intersection set is, the more probable that a false match will fall into it. Fig. \ref{fig:inter_area_distribution} depicts the distribution of this cardinality ratio on different database scales.

\textbf{Estimation of term 2} In contrast to term 1, the probability encoded in term 2 reflects the likelihood that $y$, a true neighbor of query $x$, falls into the intersection set $\mathcal{A} \cap \mathcal{B}$.

Still, we estimate this probability as a function of the cardinality ratio $\frac{Card(\mathcal{A} \cap \mathcal{B})}{Card(\mathcal{A}\cup \mathcal{B})}$. However, since the number of true matches is very small compared to false ones, we do not adopt the method in estimating term 1. Instead, image data with ground truth is used to analyze the distribution.

\begin{figure}
  \centering
  \includegraphics[width=3.1in]{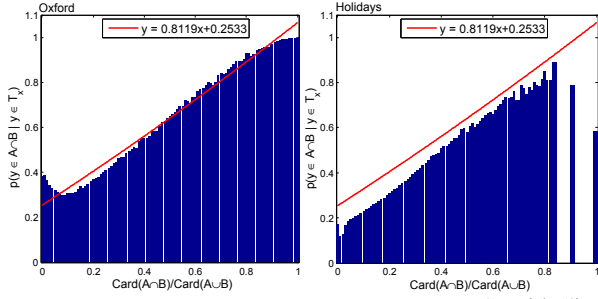}
  \caption{Distribution of term 2 as a function of $\frac{Card(\mathcal{A} \cap \mathcal{B})}{Card(\mathcal{A} \cup \mathcal{B})}$ on Oxford (\textbf{left}) and Holidays (\textbf{right}) datasets. Least Square Fitting of degree $1$ is performed on Oxford, plotted as the red line. We find that the same line also fits the trends of Holidays dataset.}
  \label{fig:term2}
\end{figure}

Specifically, empirical analysis is performed on Oxford and Holidays datasets. Given a feature $x$ in the query image $Q$, true matches are defined as the features which have a Hamming distance \cite{hamming} smaller than $20$ to $x$ \emph{and} which appear in the ground truth images of $Q$. Then we calculate the ratio of the number of true matches in $\mathcal{A}\cap \mathcal{B}$ to the number of true matches in $\mathcal{A}\cup \mathcal{B}$. Finally, the relationship between the ratio and $\frac{Card(\mathcal{A} \cap \mathcal{B})}{Card(\mathcal{A} \cup \mathcal{B})}$ is depicted in Fig. \ref{fig:term2}.

A surprising fact from Fig. \ref{fig:term2} is that $p( \mathcal{A} \cap \mathcal{B}\left.\right| F_x)$ increases linearly with $\frac{Card(\mathcal{A} \cap \mathcal{B})}{Card(\mathcal{A} \cup \mathcal{B})}$. Contrary to our expectation, true matches do not aggregate around the query point. Instead, they tend to scatter in the high-dimensional feature space. Otherwise, the curves in Fig. \ref{fig:term2} would take on a $\log(\cdot)$-like profile.
On the other hand, Fig. \ref{fig:term2} also implies that the indexed features in $\mathcal{A} \cap \mathcal{B}$ are mostly false matches. This explains why the over-counting compromises the retrieval accuracy. Moreover, we also find that the trend in Fig. \ref{fig:term2} seems to be database-independent.

\textbf{Estimation of term 3} Term 3, \ie, $p(F_x) \slash p(T_x)$, can be interpreted as the ratio of the probability of $y$ being a false match to $y$ being a true match. Typically, as the database grows, the number of false images will become larger, and the value of term 3 will increase. To model this property, and thus making our system adjustable to large scale settings, we set term 3 as

\begin{equation}\label{equation:term3}
  \frac{p(F_x)}{p(T_x)} = \log\left(N\cdot c\right),
\end{equation}
where $N$ is the number of images in the database, and $c$ is a weighting parameter. Note that we add a $\log(\cdot)$ operator due to numerical considerations.

\subsection{Similarity Interpretation}
\label{section: similarity}
\begin{figure}
  \centering
  \includegraphics[width=2.2in]{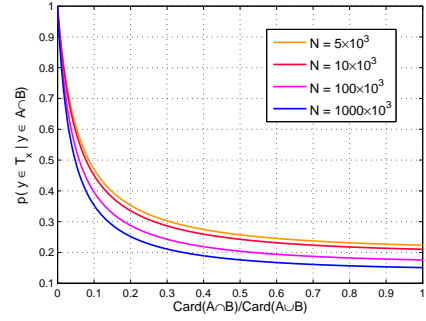}\\
  \caption{The estimation of Eq. \ref{equation: conditional_probability} as a function of $\frac{Card(\mathcal{A} \cap \mathcal{B})}{Card(\mathcal{A} \cup \mathcal{B})}$. Four curves are presented, corresponding to $N = $ 5K, 10K, 100K, and 1M, respectively. The vocabulary sizes are both 20K.}
  \label{fig:function}
\end{figure}

Using the estimation methods introduced in Section. \ref{section: prob_estimation}, we are able to provide an explicit implementation of the probability model (Eq. \ref{equation: conditional_probability}). Specifically, we assume four database sizes are involved, \ie, 5K, 10K, 100K, 1M, and we set the parameter $c$ to $1$ for better illustration. The derived probability function is plotted against $\frac{Card(\mathcal{A} \cap \mathcal{B})}{Card(\mathcal{A} \cup \mathcal{B})}$ in Fig. \ref{fig:function}. From the curves in Fig. \ref{fig:function}, we can get several implications in terms of physical interpretation.

First, when the intersection area is very small (the cardinality ratio is close to zero), it is very likely that $y$ is a true match if it falls into this area. In this scenario, the discriminative power of the intersection set is high, and can be trusted when merging vocabularies.

Second, when the cardinality ratio approaches $1$, \ie, sets $\mathcal{A}$ and $\mathcal{B}$ share a large overlap, the probability of $y$ being a true match is small. This makes more sense if we take into consideration the fact that false images dominate the entire feature space. Moreover, a larger intersection means a larger dependency (or correlation) between two vocabularies, in which situation our method exerts a punishment (low weight) and overcomes this problem to some extent.

Third, as the database becomes larger, the curves lean towards the origin. In fact, for large databases, the chances that $y$ is a true match will be more remote under each cardinality ratio. Nevertheless, the cardinality ratio tends to get smaller (see Fig. \ref{fig:inter_area_distribution}) as the database grows, so the estimated probability will be compensated to some extent.

As a summary, Fig. \ref{fig:function} reveals some interesting properties of $\mathcal{A} \cap \mathcal{B}$. The formula Eq. \ref{equation: conditional_probability} will be adopted into the BoW-based image retrieval framework in Section. \ref{section: retrieval_pipeline}.
\subsection{Generalization to Multiple Vocabularies}
\label{section:generalization}
In this section, we generalize our method to the case of multiple vocabularies ($K\geq2$).

Given $K$ vocabularies, a query feature $x$ is quantized to $K$ visual words, and subsequently $K$ sets of indexed features are identified, \ie, $\{\mathcal{A}_i\}_{i = 1}^K$. If a database feature $y$ falls into the $k^{th}$-order intersection set of $\{\mathcal{A}_i\}_{i = 1}^K$, the probability of it being a true match to $x$ is defined as
\begin{equation}\label{equation: conditional_probability_multiple}
  w(x, y) = p(y \in T_x \left.\right| y \in  \mathcal{A}_1 \cap \mathcal{A}_2 \cap ... \cap \mathcal{A}_k).
\end{equation}
Using the similarity function derived in Section \ref{section: similarity}, we can estimate Eq. \ref{equation: conditional_probability_multiple} as a function of the cardinality ratio of the $k^{th}$-order intersection and union sets $ \frac{Card\left(\mathcal{A}_1 \cap \mathcal{A}_2 \cap ... \cap \mathcal{A}_k \right)}{ Card\left(\mathcal{A}_1 \cup \mathcal{A}_2 \cup ... \cup \mathcal{A}_k\right)}$.

\subsection{Proposed Image Retrieval Pipeline}
\label{section: retrieval_pipeline}

\begin{algorithm}[t]         
\caption{ Bayes merging for image retrieval}             
\label{alg:Framwork}                  
\begin{algorithmic}[1]                
\REQUIRE ~~\\                          
    The query image $Q$ with $L$ descriptors $x_1, x_2,..., x_L$;\\
    The $K$ vocabularies $\mathcal{V}^{(1)}, \mathcal{V}^{(2)},..., \mathcal{V}^{(K)}$;\\
    The $K$ inverted files $\mathcal{W}^{(1)}, \mathcal{W}^{(2)},..., \mathcal{W}^{(K)}$;
\ENSURE ~~\\
\FOR {$n=1: L$}
\STATE Quantize $x_n$ into $K$ visual words $v^{(1)},...,v^{(K)}$; \label{code:quantization}
\STATE Identify $K$ lists of indexed features $\mathcal{A}^{(1)},...,\mathcal{A}^{(K)}$; \label{code:indexed_features}
\STATE Find all $k^{th}$-order intersection sets, $k = 2,...,K$;\label{code:find_inter_sets}
\STATE Find all $k^{th}$-order union sets, $k = 2,...,K$;\label{code:find_union_sets}
\FOR {\mbox{each indexed feature in $\cup^K$}}
\STATE Find the $k^{th}$-order intersection set it falls in; \label{code:find_one_set}
\STATE Find the $k^{th}$-order union set it falls in; \label{code:find_union_set}
\STATE Calculate $\frac{Card\left(\cap^{(k)}\right)}{Card\left(\cup^{(k)}\right)}$;\label{code:cal_ratio}
\STATE Calculate matching strength using Eq. \ref{equation: conditional_probability_multiple};\label{code:cal_strength}
\STATE Vote for the candidate image using Eq. \ref{equation:matching_multiple_ours};\label{code:vote}
\ENDFOR
\ENDFOR                       

\end{algorithmic}
\end{algorithm}


In this section, the matching function of the Bayes merging method is defined as follows,
\begin{equation}\label{equation:matching_multiple_ours}
f(x, y) =
\begin{cases}
kw(x, y), &\mbox{if } y \in \cap^k, k\geq2\\
\sum_{i=1}^{K}\delta_{v_x^{(k)},  v_y^{(k)}}, &\mbox{otherwise}
\end{cases}
\end{equation}
where $w(x, y)$ is the similarity function defined in Eq. \ref{equation: conditional_probability_multiple}. If $w(x, y) = 1$, Bayes merging reduces to the baseline B$_1$.

The pipeline of Bayes merging is summarized in Algorithm \ref{alg:Framwork}. In the offline steps, $K$ vocabularies are trained and the corresponding $K$ inverted files are organized. During online retrieval, given a query image $Q$ with $L$ descriptors, for each feature $x_n$, we quantize it to $K$ visual words (step \ref{code:quantization}). Then, $K$ lists of indexed features are identified (step \ref{code:indexed_features}), from which all $k^{th}$-order intersection and union sets are identified (step \ref{code:find_inter_sets}, \ref{code:find_union_sets}). For each indexed feature in $\cup^K$, we find the $k^{th}$-order intersection and union sets it falls in (step \ref{code:find_one_set}, \ref{code:find_union_set}), and calculate the cardinality ratio (step \ref{code:cal_ratio}). Finally, matching strength is calculated according to Eq. \ref{equation: conditional_probability_multiple} and used in the matching function as Eq. \ref{equation:matching_multiple_ours} (steps \ref{code:cal_strength} and \ref{code:vote}).

For one query feature, we have to traverse $\cup^K$ twice in Algorithm \ref{alg:Framwork}, which doubles the query time. However, in the \textbf{supplementary material}, we demonstrate that we can accomplish this process by traversing $\cup^K$ only once, thus solving the efficiency problem of Bayes merging.
\section{Experiments}
\label{section:experiments}
In this section, the proposed Bayes merging is evaluated on three benchmark datasets, \ie, Holidays \cite{hamming}, Oxford \cite{AKM}, and Ukbench \cite{HKM}. The details of the datasets are summarized in Table \ref{table:datasets}. We also add the Flickr 1M dataset \cite{hamming} of one million images to test the scalability of our method. All the vocabularies are trained independently on the Flickr60K dataset \cite{hamming} using AKM \cite{AKM} with different initial seeds.

\setlength{\tabcolsep}{2.7pt}
\begin{table}[t]
\renewcommand{\arraystretch}{1.2}
\centering
\begin{tabular}{|l|c|c|c|c|}
\hline
Dataset& \# images &  \# queries & \# descriptors &  Evaluation \\

\hline
\hline
\emph{Holidays}& 1491 & 500 & 4,455,091 & mAP \\
\hline
\emph{Oxford}& 5063 & 55 & 13,516,660 & mAP\\
\hline
\emph{Ukbench}& 10200 & 10200 & 19,415,079 & N-S score\\
\hline
\end{tabular}
\caption{Details of the datasets in the experiments.}
\label{table:datasets}
\end{table}

%
%
%

\subsection{Parameter Analysis}
\label{section: parameter}
One parameter, \ie, the weighting parameter $c$ in Eq. \ref{equation:term3} is involved in the probabilistic model. We evaluate $c$ on the Holidays and Oxford datasets, and record in Table \ref{table:parameter} the mAP results against different values of $c$. We can see that the mAP results remain stable when $c$ ranges from $10$ to $50$, probably due to the effect of the \emph{log} operator in Eq. \ref{equation:term3}. We therefore set $c$ to $30$ in the following experiments.

\setlength{\tabcolsep}{3.5pt}
\begin{table}[t]
\renewcommand{\arraystretch}{1.2}
\centering
\begin{tabular}{|l|c|c|c|c|c|}
\hline
Value of $c$ in Eq. \ref{equation:term3}& 10 &  20 & 30 &  40 & 50 \\

\hline
\hline
\emph{Oxford}, mAP(\%)& 46.72 & 46.80 & 46.77 & 46.73 &  46.73  \\
\hline
\emph{Holidays}, mAP(\%)& 58.35 & 58.47 & 58.51 & 58.50 & 58.36\\
\hline
\end{tabular}
\caption{The impact of parameter $c$ on image retrieval accuracy. Results (mAP in percent) on Oxford 5K and Holidays datasets are presented. We set $c = 30$ from these results.}
\label{table:parameter}
\end{table}

\subsection{Evaluation}
\begin{figure}
  \centering
  \includegraphics[width=3.30in]{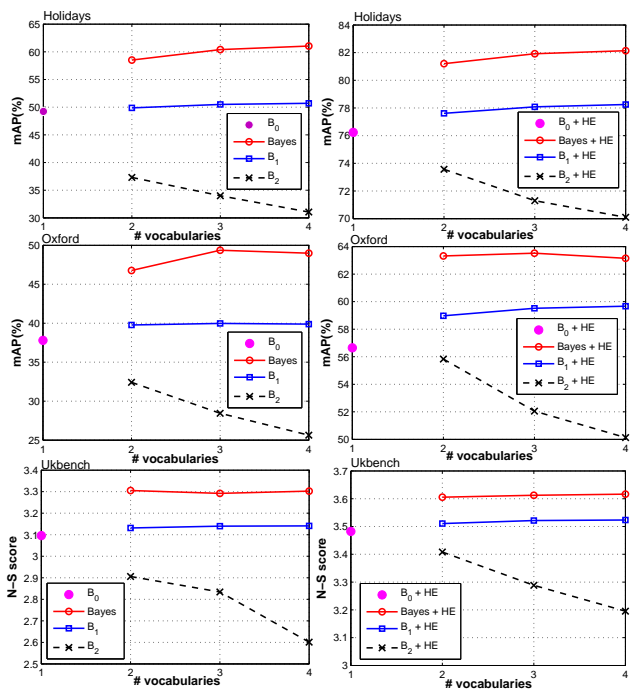}\\
  \caption{Image retrieval accuracy as a function of the number of merged vocabularies. Results of Holidays (\textbf{Top}), Oxford (\textbf{Middle}), and Ukbench (\textbf{Bottom}) are presented. We compare three baselines, \ie, $B_0$, $B_1$, and $B_2$ (see Section. \ref{section: baseline}), with our method (Bayes) (\textbf{Left}). We also show the results combined with Hamming Embedding (HE) (\textbf{Right}). The vocabulary size is 20K.    }\label{fig: HE_baselines}
\end{figure}

\makeatother
\begin{figure*} [t]
\centering
\subfigure[Holidays]{\label{fig:holidays_voc_size}%
\includegraphics[width=2.2in]{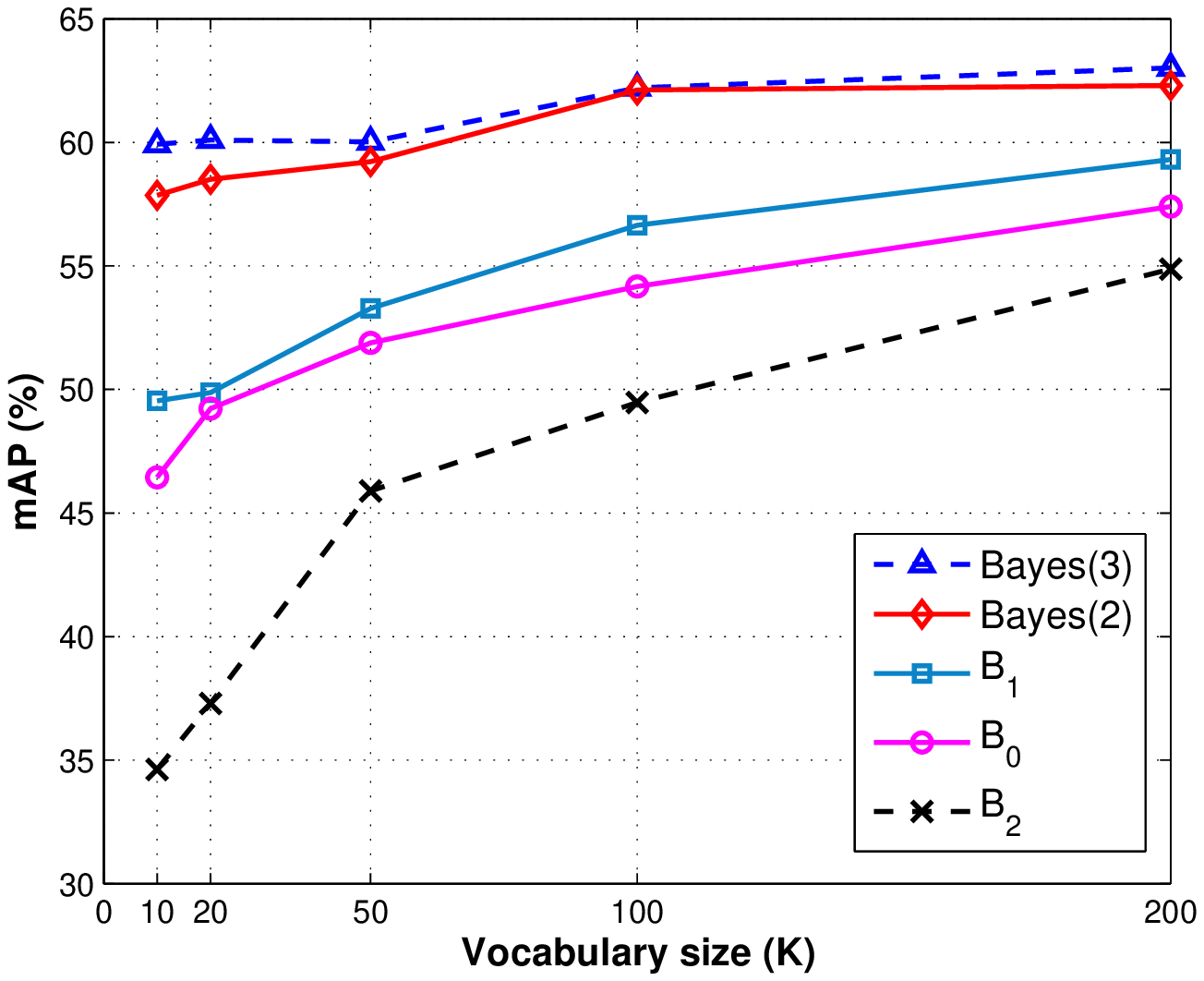}}
\hspace{0.in}
 \subfigure[Oxford]{\label{fig:oxford_voc_size}%
\includegraphics[width=2.2in]{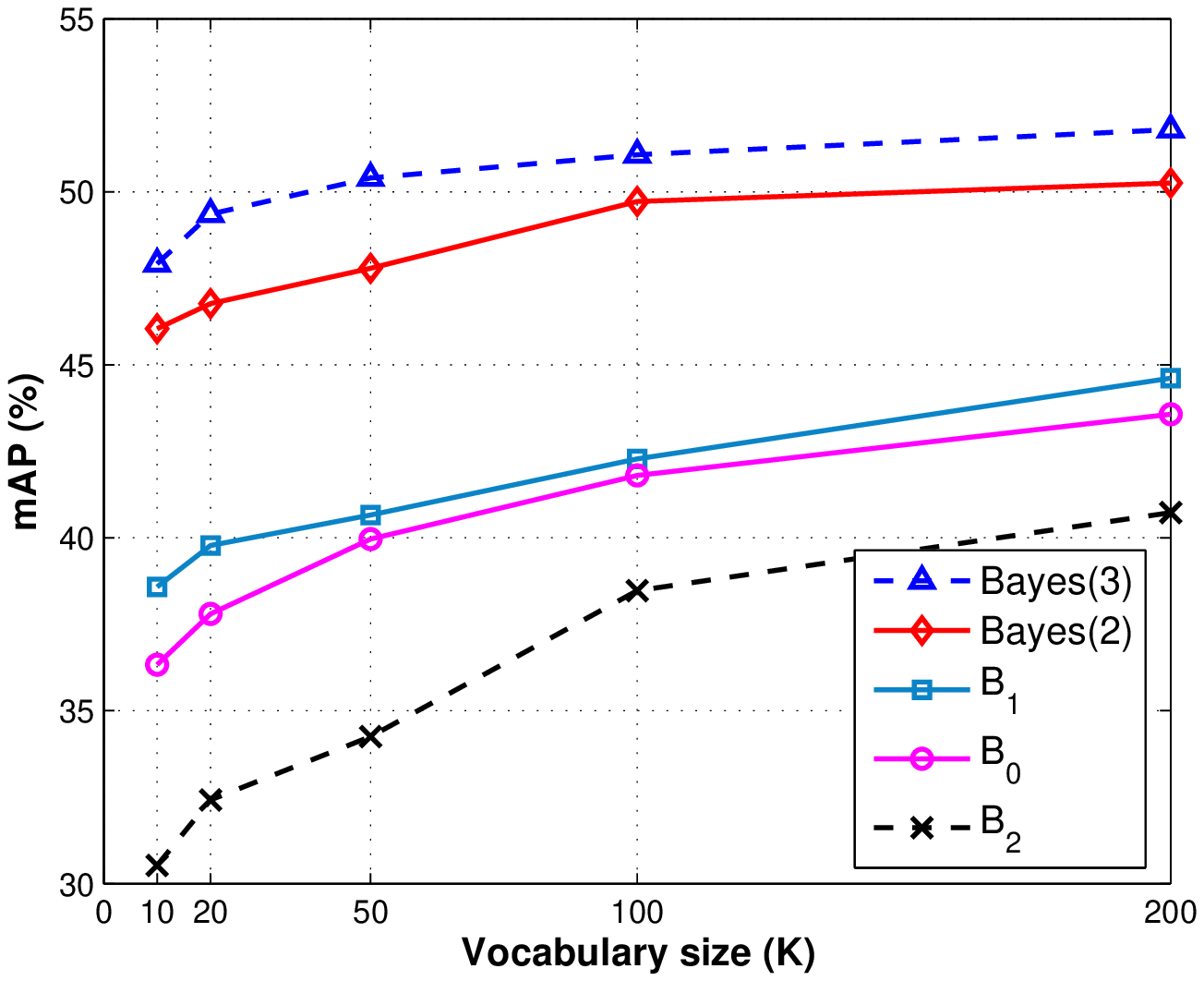}}
\hspace{0.in}
 \subfigure[Ukbench]{\label{fig:ukbench_voc_size}%
\includegraphics[width=2.2in]{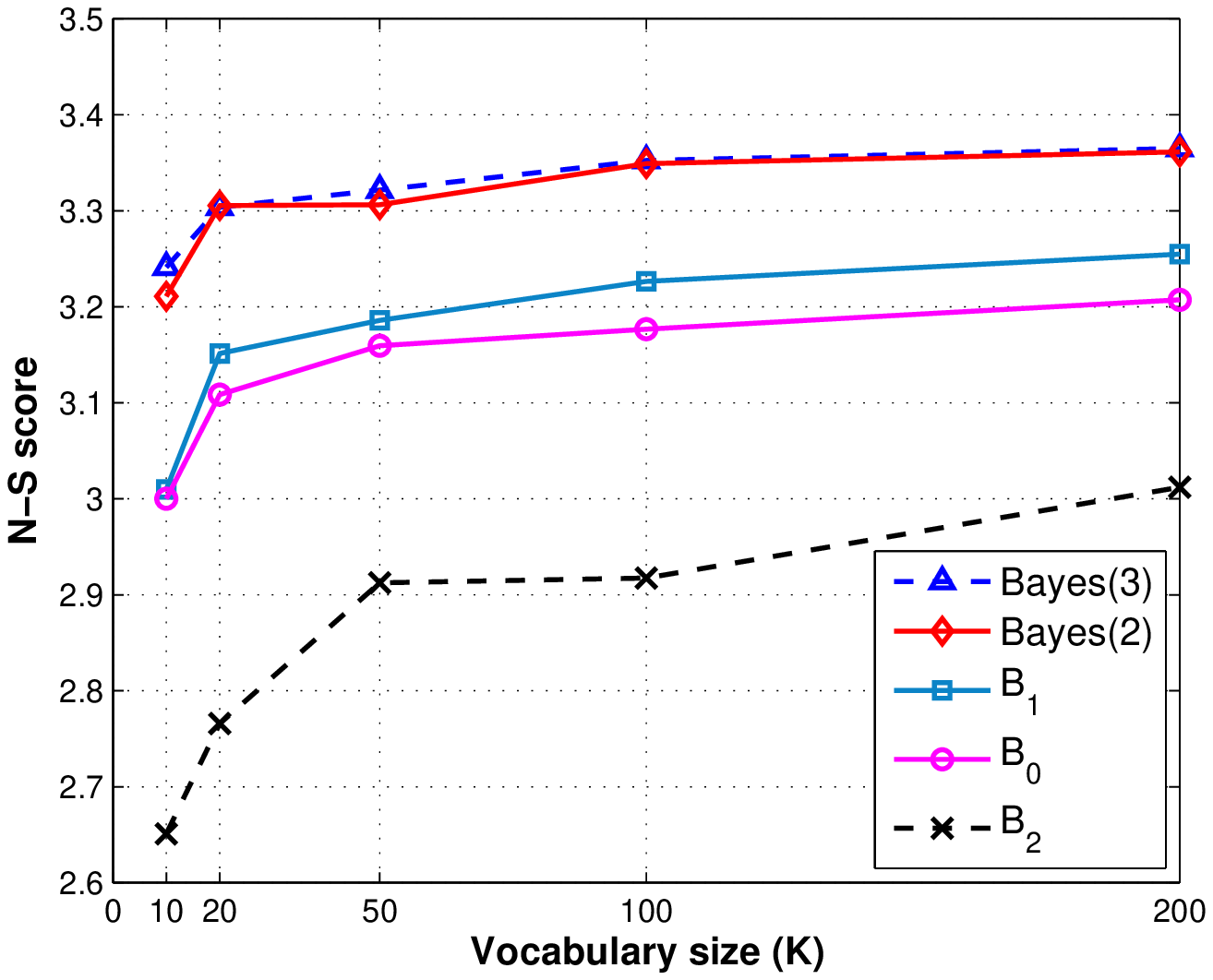}}
\caption{Image retrieval performance as a function of the vocabulary size. Methods include the three baselines, \ie, $\mbox{\bf B}_{\bf 0}$, $\mbox{\bf B}_{\bf 1}$, $\mbox{\bf B}_{\bf 2}$, as well as the proposed Bayes merging of two and three vocabularies, \ie, \textbf{Bayes(2)} and  \textbf{Bayes(3)}, respectively. mAP on (a) Holidays and (b) Oxford, and N-S score on (c) Ukbench are presented.}

\label{fig:voc_size}
\end {figure*}

\textbf{Comparison with the baselines}
We first compare Bayes merging with the baselines, \ie, $\mbox{B}_0$, $\mbox{B}_1$, $\mbox{B}_2$ defined in Section \ref{section: baseline}. The results are demonstrated in Fig. \ref{fig: HE_baselines} and Fig. \ref{fig:voc_size}. From these results we find that baseline $\mbox{B}_2$ does not benefit from introducing multiple vocabularies, and that its performance drops when merging more vocabularies, because the recall further decreases. We speculate that Multiple Assignment will bring benefit \cite{babenko2012inverted, wu2009multi} to B$_2$. Moreover, baseline B$_1$ brings limited improvements over B$_0$. In fact, B$_1$ has a higher recall than B$_0$, but this benefit is impaired by vocabulary correlation in which many irrelevant images are over-counted.


In comparison, it is clear that Bayes merging yields great improvements. Take Holidays for example, when merging two vocabularies of size 20K, the gains in mAP over the three baselines are $9.28\%$, $8.64\%$, and $21.21\%$, respectively. The improvement is even higher for three vocabularies. Nevertheless, we favor two vocabularies due to the fact that the marginal improvement is prominent, while introducing little computational complexity.

\setlength{\tabcolsep}{6.7pt}
\begin{table*}[!t]
\renewcommand{\arraystretch}{1}
\begin{center}
\begin{tabular}{|l|ccc|ccc|ccc|}
\hline
\multirow{2}{*}{Methods} &
\multicolumn{3}{c|}{\emph{Holidays}, mAP($\%$)} &
\multicolumn{3}{c|}{\emph{Oxford}, mAP($\%$)} &
\multicolumn{3}{c|}{\emph{Ukbench}, N-S}\\

\cline{2-10}

& 2${\times}$20K & 3${\times}$20K & 2${\times}$50K & 2${\times}$20K &  3${\times}$20K & 2${\times}$100K &  2${\times}$20K & 3${\times}$20K & 2${\times}$100K\\
\hline
\hline
$\mbox{B}_0$ & 49.23 &   49.23   & 54.17 & 37.80 &  37.80    &  38.80  &  3.11 &  3.11  &  3.17 \\
\hline
$\mbox{B}_1$ & 49.87 &   50.49 & 56.64 & 39.78  & 39.97 &  41.28  & 3.15  & 3.16 & 3.22\\
\hline
    Bayes     & 58.51 &   60.40 & 59.22 & 46.77  & 49.36 &  49.72  & 3.31 & 3.30 & 3.35\\
\hline
$\mbox{B}_1$ + HE & 77.61 & 78.08 & 77.48 & 58.97 & 59.52 & 60.08 & 3.51 & 3.51 & 3.48\\
\hline
    Bayes + HE&  81.20&   81.56 &    80.60   & 63.32  & 63.53 & 63.96  & 3.61  & 3.62 & 3.57\\
\hline
    Bayes + HE + Burst & 81.53 & $\bf 81.92$ & 81.08 & \bf 65.01  &  64.82 &  64.73 & \bf 3.62 & 3.62 & 3.59\\
\hline

\end{tabular}
\caption{Results on three benchmark datasets for different methods: baselines $\mbox{B}_0$ and $\mbox{B}_1$, the proposed method (Bayes), Hamming Embedding (HE) \cite{hamming}, and burstiness weighting (Burst) \cite{burstiness}. We consider the merging of 2$\times$20K, 3$\times$20K, and 2$\times$50K vocabularies, respectively.}
\label{table:various_approaches}
\end{center}
\end{table*}
\setlength{\tabcolsep}{1.4pt}

\textbf{Impact of vocabulary sizes}
The vocabulary size may have an impact on the effectiveness of Bayes merging. To this end, we generate vocabularies of size 10K, 20K, 50K, 100K, and 200K on the independent Flickr60K data. In Fig. \ref{fig:voc_size}, we demonstrate the results obtained from various vocabulary sizes on the three datasets. Except for the three baselines, we also report results obtained by Bayes merging of two or three vocabularies.

From Fig. \ref{fig:voc_size}, we can see that B$_1$ still yields limited improvement over B$_0$. Moreover, B$_1$ and B$_2$ perform better under those larger vocabularies. This is due to the fact that larger vocabularies reduce correlation. But for large databases, vocabularies are never large enough, so the correlation problem would be more severe in the large-scale case. Moreover, it is clear that the Bayes merging method exceeds the baselines consistently under different vocabulary sizes. Meanwhile, Bayes merging of three vocabularies has a slightly higher performance than two vocabularies.
%
%
%

\textbf{Merging vocabularies of different sizes}
Bayes merging can also be generalized to merging vocabularies of different sizes, and the procedure is essentially the same with Algorithm \ref{alg:Framwork}. As with the contribution of each vocabulary, we adopt the same unit weight for all vocabularies, as it is shown to yield satisfying performance in \cite{jegou2012negative}. In this paper, we report the merging results on Oxford dataset in Table \ref{table:merging_voc_different_size}.

Table \ref{table:merging_voc_different_size} demonstrates that merging vocabularies of different sizes marginally improves mAP on Oxford. For example, Bayes merging of two vocabularies of size 10K and 20K improves over the 2$\times$10K and 2$\times$20K Bayes methods by $1.07\%$ and $0.34\%$, respectively. We speculate that vocabularies of different sizes provide extra complementary information, which can be captured by our method. However, since the smaller vocabulary introduces more noise, the benefit is limited.

\setlength{\tabcolsep}{11pt}
\begin{table}[t]
\renewcommand{\arraystretch}{1.0}
\centering
\begin{tabular}{|l|c|c|c|}
\hline
Method& $\mbox{B}_1$ &  $\mbox{B}_2$ & Bayes\\

\hline
\hline
10K + 20K& 40.89 & 32.85 & 47.11  \\
\hline
20K + 50K& 41.20 & 34.70 & 48.85 \\
\hline
10K + 20K + 50K& 42.31 & 35.82 & 49.05\\
\hline
\end{tabular}
\caption{The mAP of Bayes merging of vocabularies of different sizes on Oxford dataset. In comparison, Bayes merging of two vocabularies of the same size yields an mAP of $46.04\%$, $46.77\%$, and $47.79\%$ for the 10K, 20K, 50K vocabularies, respectively.}
\label{table:merging_voc_different_size}
\end{table}

\textbf{Combination with Hamming Embedding} To test whether Bayes merging is complementary to some prior arts, we combine it with Hamming Embedding (HE) \cite{hamming} and burstiness weighting \cite{burstiness} using the default parameters. HE effectively improves the precision of feature matching. In our experiment, HE with a single vocabulary achieves an mAP of $76.24\%$ and $56.65\%$ on Holidays and Oxford, and an N-S score of $3.49$ on Ukbench, respectively.

The results in Fig. \ref{fig: HE_baselines} and Table \ref{table:various_approaches} indicate that Bayes merging yields consistent improvements of the B$_0$ + HE method. Specifically, when merging two vocabularies of 20K, the mAP is improved from $76.24\%$ to $81.20\%$ and from $56.65\%$ to $63.32\%$ on Holidays and Oxford, respectively. Similar trend can be observed on Ukbench: N-S score rises from 3.49 to 3.61. In its nature, HE results in refined matching in the feature space (locally). Complementarily, the Bayes merging jointly considers the image- and feature-level similarity. Therefore, while good matching in the feature space can be guaranteed by HE, our method punishes those of a false match in the image space. In this scenario, we actually raise an interesting question: \emph{can we simply trust feature-to-feature similarity in image retrieval?}

In addition, combining burstiness weights brings about extra, though limited improvement (see Table \ref{table:various_approaches}). Our implementation differs from \cite{burstiness} in that we do not apply the weights on images in the intersection set, but instead on the difference set ($\mathcal{A}\cup \mathcal{B} - \mathcal{A}\cap \mathcal{B}$) only. A performance summary of various methods is presented in Table \ref{table:various_approaches}.

\begin{figure}
  \centering
  \includegraphics[width=3.30in]{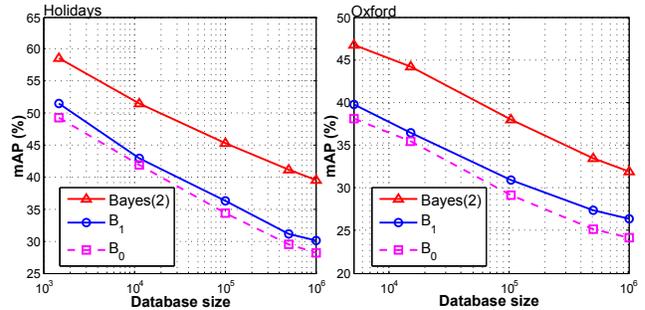}\\
  \caption{The mAP results as a function of the database size on Holidays and Oxford datasets. Three methods are compared, \ie, baselines B$_0$, B$_1$, and Bayes merging of two vocabularies. The vocabulary size is 20K for all methods.}\label{fig:large_scale}
\end{figure}

\textbf{Large-scale experiments}
To test the scalability of our method, we add the Flickr1M distractor images \cite{hamming} to the Holidays and Oxford datasets. For comparison, we report the results of baselines $\mbox{B}_0$ and $\mbox{B}_1$. From Fig. \ref{fig:large_scale}, it is clear that Bayes merging outperforms the two baselines significantly. On Holidays dataset mixed with one million images, Bayes merging achieves mAP of 39.60$\%$, compared with 28.19$\%$ and 29.26$\%$ of baseline B$_0$ and B$_1$, respectively.

In terms of efficiency, the baseline method  $\mbox{B}_0$ consumes 4 bytes per feature, and 1.9 GB for indexing one million images. The Bayes merging of two vocabularies doubles the memory cost to about 3.8 GB on Flickr1M.

On the other hand, it takes 2.52s and 4.87s for $\mbox{B}_0$ and $\mbox{B}_1$ to perform one query on 1 million image size, respectively, using a server with 3.46 GHz and 64GB memory. Bayes merging involves identifying the intersection set and calculate the cardinality ratio. In fact, the cardinality ratio can be computed and stored offline. Moreover, as shown in the \textbf{supplementary material}, we are able to perform both the identification and the voting tasks by traversing the two lists of indexed features only once. Therefore, our method only marginally increases the query time to 5.12s.

\textbf{Comparison with state-of-the-arts}
We first compare our method with \cite{jegou2012negative} which employs PCA to addresses the correlation problem implicitly. In \cite{jegou2012negative}, merging four 16K vocabularies and eight 8K vocabularies yield an mAP of $55.8\%$ and $56.7\%$, respectively. Moreover, merging vocabularies of multiple sizes obtains a best mAP of $58.8\%$ on Holidays. In comparison, the result obtained by Bayes merging is $58.5\%$ and $60.4\%$ for two and three vocabularies of size 20K, respectively.

Second, we compare the Bayes merging with the Rank Aggregation (RA) method \cite{fagin2003efficient, jegou2010accurate} in Table \ref{table:comp_rank_agg}. Following \cite{jegou2010accurate}, we take the median of multiple ranks as the final rank. Since RA works on the rank level, it does not address the correlation problem, so its performance is limited. The results demonstrate the superiority of Bayes merging.

Finally, we compare the results of Bayes merging with state-of-the-arts in Table \ref{table:state_of_art}. On the three datasets, we achive mAP = $\bf{81.9\%}$ on Holidays, mAP = $\bf{65.0\%}$ on Oxford, and N-S = $\bf3.62$ on Ukbench. We have also tested on the data provided by \cite{selective_match}, where the codebook size is 65K. On Oxford datastet, the mAP is 77.3\%. Note that some sophisticated techniques are absent in our system, such as spatial constraints \cite{jegou2010improving, contextual_weighting}, semantic consistency \cite{co_indexing}, etc. Still, the results demonstrate that the performance of Bayes merging is very competitive. We also provide some sample retrieval results in the \textbf{supplementary material}.

\setlength{\tabcolsep}{3.4pt}
\begin{table}[t]
\renewcommand{\arraystretch}{1.0}
\centering
\begin{tabular}{|l|cc|cccc|}
\hline
Method&
\multicolumn{2}{c|}{Bayes}  &
\multicolumn{4}{c|}{Rank aggregation}\\
\hline
\# vocabularies& 2 & 3 & 3 & 5 & 7 & 9\\
\hline
\hline
\emph{Holidays},  mAP(\%) & 58.5 & 60.4 & 49.9  & 50.7 & 51.0 & 51.0 \\
\hline
\emph{Oxford},  mAP(\%) & 46.8 & 49.4 &  39.5 &  40.8 & 41.2 & 41.4 \\
\hline
\end{tabular}
\caption{Comparisons with rank aggregation (RA). Different numbers of vocabularies are trained to test RA. Vocabulary size is 20K.}
\label{table:comp_rank_agg}
\end{table}

\setlength{\tabcolsep}{3.3pt}
\begin{table}[t]
\renewcommand{\arraystretch}{1.0}
\centering
\begin{tabular}{|l|cccccc|}
\hline
 Method& Bayes &  \cite{co_indexing} & \cite{jegou2010improving} & \cite{burstiness} &\cite{contextual_weighting} & \cite{wang2012fast}\\

\hline
\hline
\emph{Holidays},  mAP(\%) &  81.9 & 80.9 & 81.3  & 83.9 & 78.1 &-\\
\hline
\emph{Oxford}, mAP(\%)&  65.0 & 68.7 & 61.5 & 64.7 & -& 66.4 \\
\hline
\emph{Ukbench} , N-S& 3.62 & 3.60 &  3.42 & 3.54 & 3.56 & 3.50\\
\hline
\end{tabular}
\caption{Comparisons with the state-of-the-art methods.}
\label{table:state_of_art}
\end{table}

\section{Conclusion}
\label{section:conclusion}
 Multi-vocabulary merging is an effective method to improve the recall of visual matching. However, this process is impaired by vocabulary correlation. To address the problem, this paper proposes a Bayes merging approach to explicitly estimate the matching strength of the indexed features in the intersection sets, while preserving those in the difference set. In a probabilistic view, Bayes merging is capable of jointly modeling an image- \emph{and} feature-level similarity from multiple sets of indexed features. Specifically, we exploit the probability that an indexed feature is a true match (both locally and globally) if it is located in the intersection sets of multiple inverted files. Extensive experiments demonstrate that Bayes merging effectively reduces the impact of vocabulary correlation, thus improving the retrieval accuracy significantly. Further, our method is efficient, and yields competitive results with state-of-the-arts.\\

\noindent\textbf{Acknowledgement} This work was supported by the National High Technology Research and Development Program of China (863 program) under Grant No. 2012AA011004 and the National Science and Technology Support Program under Grant No. 2013BAK02B04. This work also was supported in part to Dr. Qi Tian by ARO grant W911NF-12-1-0057, Faculty Research Awards by NEC Laboratories of America,  and 2012 UTSA START-R  Research Award respectively. This work was supported in part by National Science Foundation of China (NSFC) 61128007.


{\footnotesize
\bibliographystyle{ieee}
\bibliography{egbib}
}

\end{document}